\documentclass{bmvc2k}

\usepackage{microtype}
\usepackage{graphicx}
\usepackage{subfigure}
\usepackage{booktabs} 

\usepackage{hyperref}
\usepackage[capitalize,noabbrev]{cleveref}

\usepackage{algorithm, algorithmic}

\usepackage{amsmath}
\usepackage{amssymb}
\usepackage{mathtools}
\usepackage{amsthm}
\usepackage{multirow}
\newcommand{\e}[1]{\ensuremath{10^{#1}}}
\usepackage{colortbl}
\usepackage{xcolor}
\usepackage[symbol]{footmisc}

\title{SCAAT: Improving Neural Network Interpretability via Saliency Constrained Adaptive Adversarial Training}

\addauthor{Rui Xu}{xurui@stu.pku.edu.cn}{$\ast$1}
\addauthor{Wenkang Qin}{qinwk@stu.pku.edu.cn}{$\ast$1}
\addauthor{Peixiang Huang}{huangpx@stu.pku.edu.cn}{1,3}
\addauthor{Hao Wang}{wanghao@nifdc.org.cn}{2}
\addauthor{Lin Luo}{luol@pku.edu.cn}{$\dag$1}
\addinstitution{
 College of Engineering\\
 Peking University\\
 Beijing, China
}
\addinstitution{
 National Institutes for Food and Drug Control\\
 Beijing, China
}
\addinstitution{
 Beijing Institute of Collaborative Innovation\\
 Beijing, China
}

\runninghead{R.Xu et al}{IMPROVING SALIENCY VIA ADAPTIVE ADVERSARIAL TRAINING}

\def\etal{\emph{et al}\bmvaOneDot}
\begin{document}

\maketitle
\renewcommand{\thefootnote}{\fnsymbol{footnote}}
\footnotetext[1]{These authors contributed equally to this work.} \footnotetext[2]{Corresponding author.}

\begin{abstract}
Deep Neural Networks (DNNs) are expected to provide explanation for users to understand their black-box predictions. Saliency map is a common form of explanation illustrating the heatmap of feature attributions, but it suffers from noise in distinguishing important features. In this paper, we propose a model-agnostic learning method called Saliency Constrained Adaptive Adversarial Training (SCAAT) to improve the quality of such DNN interpretability. By constructing adversarial samples under the guidance of saliency map, SCAAT effectively eliminates most noise and makes saliency maps sparser and more faithful without any modification to the model architecture. We apply SCAAT to multiple DNNs and evaluate the quality of the generated saliency maps on various natural and pathological image datasets. Evaluations on different domains and metrics show that SCAAT significantly improves the  interpretability of DNNs by providing more faithful saliency maps without sacrificing their predictive power. 
\end{abstract}

\section{Introduction}
\label{submission}
With the fast development of deep neural networks, model interpretability has become an essential part of building reliable and robust models in critical application domains such as pathological diagnosis\cite{ma2023agmdt, qin2022pathtr, huang2023assessing, qin2023whole}, drug discovery\cite{preuer2019interpretable}, cross-modal reasoning\cite{yang2023improving} and autonomous driving\cite{huang2022tigbev}.

Saliency methods are techniques used to analyze the contribution of input features to model predictions. In image classification, these methods can generate a heatmap, called a saliency map \cite{simonyan2013deep}, to highlight the most crucial input regions for a model's prediction. Techniques such as SmoothGrad \cite{smilkov2017smoothgrad}, Integrated Gradient \cite{sundararajan2017axiomatic}, CAM \cite{zhou2016learning}, LRP \cite{bach2015pixel}, and DeepLIFT \cite{shrikumar2017learning} are commonly used for interpreting model predictions and understanding the decision-making process of complex models. 

By analyzing saliency maps, it is possible to quantitatively determine which input regions are the most relevant to the classification result and which are not, thus to understand the decision-making process of a model. The sparsity of the saliency map is crucial, as it helps to identify the key regions without being overwhelmed by random noise. In addition to sparsity, the faithfulness of a saliency map is a measure of how accurately it reflects the salient features of the inputs, imposing additional requirements on the saliency map generation process.

Traditional learning methods, which focus on task-related objectives and prediction performance, may have limitations in interpretability. Due to the lack of constraints on the sparsity of the model's attention, a model may be sensitive to many irrelevant features, resulting in a lot of noises in the saliency map, which impacts the interpretability of the model predictions. 

In this paper, we propose a novel model-agnostic learning method called Saliency Constrained Adaptive Adversarial Training (SCAAT) which actively introduces saliency constraint to the model training process to improve the sparsity and faithfulness of the saliency maps. Our method is distinct from general adversarial training approaches as it can adaptively select critical features from saliency maps and keep them unperturbed, thereby preserving model discrimination power on clean samples and meanwhile improving the faithfulness of the saliency maps.

The contributions of our work can be summarized as follows:
\begin{itemize}
\item We propose a novel model-agnostic adaptive adversarial training framework which improves the interpretability of deep neural networks without changing the networks, and thus it can be generalized to various models and domains.

\item We develop an adaptive perturbation searching method with an adversarial objective function which can balance the optimization between the learning performance and the resilience against perturbations on irrelevant features. 

\item To our best knowledge, this is the first work that introduces adversarial training with saliency constraints to improve neural network interpretability. Experiments on both natural and pathological image datasets show that our SCAAT outperforms the state-of-the-art interpretability approaches in measures of saliency map sparsity and faithfulness, while barely sacrificing the predictive performance of the models.
\end{itemize}

\begin{figure*}[h]
    \centering
    \includegraphics[width=1.0\textwidth]{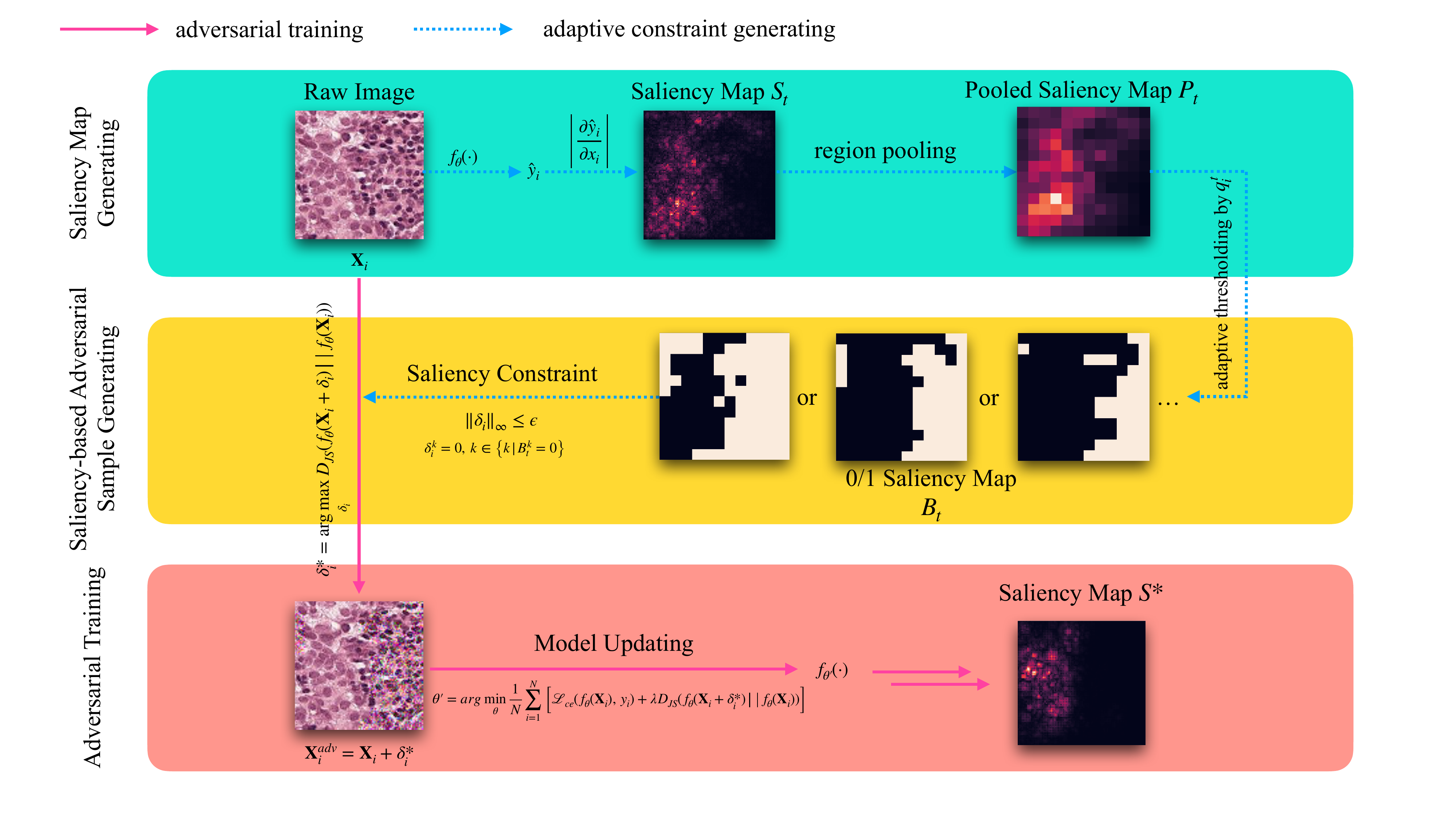}
    \caption{\textbf{An overview of our proposed SCAAT.} For each sample, we generate the region-averaged gradient-based saliency map whose resolution is same as the input image, then we select the regions to be perturbed in the adversarial training based on the proportion $q_i$. Then we involve the saliency constraints (i.e. high saliency pixels will not be perturbed) in adversarial sample generation and get the optimal adversarial sample using PGD-k \cite{madry2017towards}. We further update model to improve its robustness to perturbations on those low-saliency features and adjust $q_i$ to get more suitable feature perturbing proportion for each instance.}
\end{figure*}

\section{Related Work}
\paragraph{Interpretability}
Interpretability research is critical to deep learning and is growing rapidly.
Related work can be divided into three lines.
The first line is about post-hoc explanation methods.
Some gradient-based methods try to compute backpropagation for a modified gradient function, like \cite{baehrens2010explain,sundararajan2017axiomatic,smilkov2017smoothgrad,shrikumar2017learning,lundberg2017unified,selvaraju2017grad}.
And others \cite{zeiler2014visualizing,suresh2017clinical,ribeiro2016should,tonekaboni2020went}, called perturbation-based methods, trying to perturb areas of the input and measure how much this changes the model output.
The second line is about measure the reliability of interpretability methods \cite{adebayo2018sanity,adebayo2020debugging,ghorbani2019interpretation,hooker2019benchmark,kindermans2019reliability,petsiuk2018rise,samek2016evaluating,tomsett2020sanity}.
Other methods like \cite{ba2014deep,frosst2017distilling,ismail2019input,ross2017right,wu2018beyond} modifying neural architectures for better interpretability.
Related to our work, \cite{ghaeini2019saliency,ross2017right,ismail2021improving} incorporate explanations into the learning process.

\paragraph{Adversarial training}
Adversarial samples are perturbed samples that are usually generated by adding small perturbations to the original samples that may mislead the neural network to make erroneous predictions \cite{szegedy2013intriguing}.
There are many popular methods to generate adversarial perturbations such as FGSM \cite{goodfellow2014explaining}, PGD \cite{madry2017towards}, DeepFool \cite{moosavi2016deepfool}, FreeAT \cite{shafahi2019adversarial} and YOPO \cite{zhang2019you}. Based on these methods of adversarial sample generating, adversarial training has been widely used to make model robust to adversarial attacks \cite{pang2020bag,maini2020adversarial,schott2018towards}.
Different from those works that focus on improving the model robustness to adversarial attacks, our work aims to desensitize the model to perturbations \textit{on the irrelevant features only}.

\paragraph{Input level perturbation}
Input level perturbation during training has been previously explored.
But most of these works try to improve performance or robustness rather than interpretability.
\cite{wei2017object,kumar2017hide,li2018tell,hou2018self} use attention maps to improve segmentation performance.
And \cite{wang2019sharpen} use attention maps for training to improve performance of classfication.
\cite{devries2017improved} improve the robustness and performance for convolutional neural networks.
Related to our work, \cite{ismail2021improving} is the first work we know of improving model interpretability through input level perturbation in a self-supervised learning manner. Their work developed a pattern-fixed and interpretability-related regularization term under the guidance of saliency map, which differs from our method of generating adversarial samples systematically and adaptively to improve the model interpretability. Furthermore, our method significantly outperforms that of \cite{ismail2021improving} for both model interpretability and classification performance.

\section{Method}
\subsection{Notation}
First, let $\mathcal{D} = \left \{ \left ( \mathbf{X}_{i} \mathrm{,}\ {y_{i}} \right )  \right \}_{i = 1}^{N}$ denote the samples in training dataset, and each sample $\mathbf{X}_{i}= \left [ x_{1},...,x_{n} \right ] \in \mathbb{R}^{n}$ has $n$ features. In the classification task, the label can be formulated as $y_{i} \in \left \{  1,2,...,N_{c}\right \}$, where $N_{c}$ denotes the class number. The neural network $f_{\theta}$ with learnable parameters $\theta$ takes $\mathbf{x}=\mathbf{X}_{i} \in \mathbb{R}^{n}$ as input, and $f_{\theta}(\mathbf{x}):\mathbb{R}^{n} \longrightarrow \mathbb{R}^{N_{c}}$ denotes that the network predicts the score of $N_{c}$ classes. The supervised learning objective is minimizing the cross-entropy loss $\mathcal{L}_{ce} $ between labels and predictions, which can be formulated as follows:
\begin{equation}
    \min\limits_{\theta} \frac{1}{N} \sum\limits_{i=1}^{N} \mathcal{L}_{ce}(f_{\theta}(\mathbf{X}_{i}),\ y_{i})
\end{equation}

Assume that the model takes $\mathbf{x}=\mathbf{X}_i$ as input with a classification label $y=y_{i}$, the gradient of the confidence of class $y$ with respect to $\mathbf{x}$ is given by $\nabla _{\mathbf{x} } f_{\theta}(\mathbf{x}) \mid _y\ = \left \{g_{1},...,g_{n}  \right \} \in \mathbb{R}^n$. Let $GSmap(f_{\theta},\mathbf{x} ,y)$ denote the absolute gradient-based saliency map $\left \{\left |g_{1} \right |,...,\left |g_{n} \right | \right \}$ for sample $\mathbf{x}$ of model $f_{\theta}$.

Let $S$ be a set of $s$ real numbers $\left \{ a_1,...,a_s \right \}$, $Lowest(S, q)$ outputs a set consisted of the indexes of those elements whose value is less than the bottom $q$-quantile value in the input set $S$, $i.e.$ $Lowest(S, q) =\left \{i\ |\ a_i < Quantile(S, q) \right \}$ 

For standard adversarial training methods \cite{bai2021recent}, the most critical step is to find adversarial perturbation $\delta_{i}$ which can maximally confuse the model when being added to the clean sample $\mathbf{X}_{i}$, and the cross-entropy loss $\mathcal{L}_{ce}$ is often used as a measure of confusion. The objective of perturbation searching under an $\epsilon$-ball constraint can be formulated as follows:
\begin{equation}
    \max\limits_{\left \| \delta _i \right \|_{\infty} \le \epsilon   } \frac{1}{N} \sum\limits_{i=1}^{N} \mathcal{L}_{ce}(f_{\theta}(\mathbf{X}_{i}+\delta _{i}),\ y_{i})
\end{equation}

Given two probability distributions $P$ and $Q$ on probability space $\mathcal{X}$, the Kullback-Leibler (KL) divergence \cite{kullback1951information} from $P$ to $Q$ is defined as follows:
\begin{equation}
    D_{KL}(P\ \left |  \right |\ Q) =  {\textstyle \sum\limits_{x\in\mathcal{X} }^{}}P(x)\cdot log_2\left (\frac{P(x)}{Q(x)} \right)
\end{equation}

The Jensen–Shannon (JS) divergence \cite{lin1991divergence} $D_{JS}$ is the symmetrical form of $D_{KL}$:
\begin{equation}
    D_{JS}(P\ \left |  \right |\ Q) =  \frac{1}{2} D_{KL}(P\ \left |  \right |\ Q) + \frac{1}{2} D_{KL}(Q\ \left |  \right |\ P)
\end{equation}

\subsection{Saliency Constrained Adaptive Adversarial Training}
To improve the sparsity and faithfulness of saliency map, the noise on irrelevant features in the saliency map must be eliminated \cite{ismail2021improving}. In other words, the model prediction should be robust to the small perturbations on irrelevant features while being sensitive to critical features, which makes the saliency map clearly indicate those features that are essential in model's prediction process.
We developed a novel adversarial-based learning objective to solve this problem.

First, we search for an optimal perturbation term $\delta_i^{\ast} \in \mathbb{R}^n $ for each sample $\mathbf{X}_i$ in the training set which maximizes the JS divergence \cite{lin1991divergence} $D_{JS}$ between $f_{\theta}(\mathbf{X}_{i}+\delta _{i}^{\ast})$ and $f_{\theta}(\mathbf{X}_{i})$:

\begin{equation}
    \delta_i^{\ast} =\mathop{\arg\max}\limits_{\delta_i}   D_{JS}(f_{\theta}(\mathbf{X}_{i}+\delta _{i})\left |  \right | f_{\theta}(\mathbf{X}_{i}))
    \label{delta}
\end{equation}

Like standard adversarial training methods \cite{bai2021recent}, the feasible perturbation $\delta_i$ is restricted to a small region controlled by $\epsilon$ since arbitrary perturbations may harm the model performance on clean samples. Additionally, we involve sample-specific saliency constraint for $\delta_i$ to prevent those features with high saliency values from being perturbed, thus the saliency map will be sparser and more faithful. 
The constraint for $\delta_i$ is formulated as follows:
\begin{equation}
    \left \| \delta_i \right \|_\infty \le \epsilon
\end{equation}

and

\begin{equation}
    \delta_i^{k} = 0,\ k \notin Lowest(Smap(f_\theta,\mathbf{X}_i,y_i), \ q_i)
\end{equation}
For each sample $\mathbf{X}_i$, the value of $q_i$ determines what proportion of features in $\mathbf{X}_i$ will be perturbed, and it can be adjusted adaptively during the training process with an initialization value $q_0 \in \left [ 0,1\right]$. The maximization problem above can be effectively solved by the PGD-k algorithm \cite{madry2017towards}.

The complete learning objective for our saliency constrained adversarial training is:
\begin{equation}
\begin{aligned}
    \min\limits_{\theta}\frac{1}{N} \sum\limits_{i=1}^{N}\Big [ &\mathcal{L}_{ce}(f_{\theta}(\mathbf{X}_{i}),\ y_{i}) \\ &+ \lambda D_{JS}(f_{\theta}(\mathbf{X}_{i}+\delta _{i}^{\ast})\left |  \right | f_{\theta}(\mathbf{X}_{i}))\Big]
\end{aligned}
\end{equation}
where the $\delta_{i}^{\ast}$ is the optimal solution of the problem defined in formulation \ref{delta}, and $\lambda$ is a hyper-parameter to balance the supervised loss and the divergence loss.

\begin{algorithm}[tb]
   \caption{Saliency constrained adaptive adversarial training.}
   \label{alg1}
\begin{algorithmic}
   \STATE {\bfseries Rquire:} $N_{iter}$: Total training iterations
   
   \STATE {\bfseries Rquire:} $PGD_k(f_\theta,\mathbf{X}_i,y_i,\epsilon,IrreFeats_i)$: The k-step PGD function \cite{madry2017towards} to generate optimal adversarial perturbation $\delta_{i}^\ast$ in the constrained $\epsilon$-ball, guaranteeing $\delta_{i}^k = 0, k \notin IrreFeats_i$
   
   \STATE {\bfseries Rquire:} Feature perturbation proportions for each training sample: $q=\left \{ q_i \right \}_1^N$ initialized by $q_0 \in \left [ 0,1 \right ]$
   
   \FOR{$iter=1$ {\bfseries to} $N_{iter}$}
   \STATE Sample a batch $\left \{ \left ( \mathbf{X}_{i} \mathrm{,}\ {y_{i}} \right )  \right \}_{i = 1}^{N_{batch}}$ from the training set $\mathcal{D}$
   \FOR{$i=1$ {\bfseries to} $N_{batch}$}
   
   \STATE \textbf{Generate saliency map}
   \STATE $S_i = GSmap(f_{\theta},\mathbf{X}_i,y_i)$
   
   \STATE \textbf{Select irrelevant features}
   \STATE $IrreFeats_i = Lowest(S_i,q_i)$
   
   \STATE \textbf{Generate adversarial samples}
   \STATE $\mathbf{X}_i^{adv} = \mathbf{X}_i + PGD_k(f_\theta,\mathbf{X}_i,y_i,\epsilon,IrreFeats_i)$
   
   \STATE \textbf{Compute the loss terms}
   \STATE $L_i^{cls} = \mathcal{L}_{ce}(f_{\theta}(\mathbf{X}_{i}),\ y_{i})$
   \STATE $L_i^{adv} = D_{JS}(f_{\theta}(\mathbf{X}_{i}^{adv})\left |  \right | f_{\theta}(\mathbf{X}_{i}))$
   
   \STATE \textbf{Update $q_i$ using Algorithm \ref{alg2}}
   \ENDFOR
   
   \STATE Update the model parameters
   \STATE $\theta^{'} = arg\min\limits_{\theta}\frac{1}{N^{batch}}\sum_{i=1}^{N^{batch}}\left (L^{cls}_i + \lambda L^{adv}_i\right ) $

   \ENDFOR
   
\end{algorithmic}
\end{algorithm}

\subsection{Adaptive Feature Perturbation Proportion}
Intuitively, the ratio of irrelevant features varies across samples. For example, an image that is mostly background should have more irrelevant features than a dense one, which deserves more aggressive perturbing strategy (i.e. perturbing more low-saliency regions). 
In this sense, perturbing irrelevant features with fixed proportion for the whole dataset is sub-optimal while determining the suitable proportion for each training sample is more reasonable.

Our method optimizes perturbing proportion for each instance by adaptively adjusting the proportion $q_i$ for each sample $\mathbf{X}_i$ during training process. Before model training, the values of $\left \{ q_i \right \}_{i=1}^N$ are initialized to the same empirically selected value $q_0 \in \left [ 0,1\right ]$, and will not be adjusted in the warm-up period. After that, we reduce $q_i$ by a step of $\gamma$ if the adversarial sample $\mathbf{X}_{adv}$ generated under the constraint of $q_i$ is misclassified by the model and vice versa. We believe that if small perturbations applied to $q_i$-proportion regions has crossed the model's decision boundary and mislead the model to a wrong prediction, then the proportion should be reduced to protect model's discrimination power.

Our training method is shown in \Cref{alg1}, and the details about how we update $q_i$ are in \Cref{alg2}.
\begin{algorithm}[tb]
   \caption{ $q$ updating algorithm }
   \label{alg2}
\begin{algorithmic}
   \STATE {\bfseries Rquire:} $i$: Index of sample, $iter$: Iteration index; $N^{warm-up}$: Warm-up iterations; 
   \STATE {\bfseries Rquire:} $q_{max}, q_{min}$: Boundary values for $q$; $\gamma$: Discretization for $q$ searching.
   \IF{$iter \le N^{warm-up}$}
   \STATE Set $q_i{}'  = q_i$
   \ELSIF{$f_{\theta}(X_i^{adv})$ predicts as $y_i$}
   \STATE Set $q_i{}'  = q_i + \gamma$
   \ELSE
   \STATE Set $q_i{}'  = q_i - \gamma$
   \ENDIF
   \STATE Set $q_i{}'' = \min(\max(q_i{}',\ q_{min}),\ q_{max})$ 
   \STATE \textbf{return} $q_i{}''$
\end{algorithmic}
\end{algorithm}

\section{Experiments}
\subsection{Datasets}
To demonstrate our method can improve model interpretability across domains, we conduct experiments on the PCAM \cite{pcam}, CIFAR-10 \cite{cifar} and ImageNet-1k \cite{deng2009imagenet} dataset. PCAM \cite{pcam} is a dataset of pathological images which consists of 327680 color images ($96\times96$ px) extracted from histopathologic scans of lymph node sections and has been generally used in the domain of computational pathology.

For each dataset, we train the model by SCAAT on ResNet-18 \cite{resnet} and VGG-16 \cite{vggnet} then compare model interpretability and performance for regular training, previous method proposed by Ismail \etal \cite{ismail2021improving} and our SCAAT. 



\subsection{Quality Evaluation for Saliency Map}
\textbf{Sparsity}~
Generally, a saliency map consists of pixel-wise scores that indicate relevant pixels for model decision. Good saliency maps should highlight relevant regions only, while sub-optimal saliency maps may have much noise and lack of sparsity. We select two metrics to evaluate the sparsity of a saliency map. One is entropy, and the other is compressed saliency map size in Kbyte. 
\textbf{Faithfulness}~
Faithfulness is a measure that quantifies the extent to which the regions highlighted by a saliency map align with the true important regions for the given prediction. Compared with sparsity, faithfulness is a more comprehensive and important evaluation metric, because sparsity only evaluates the noise level of the saliency map, without considering how accurately the highlighted regions in the saliency map match the critical regions.

By following \cite{AOPC,sanitychecker}, we use AOPC to evaluate the saliency map faithfulness. Given a saliency map of a test sample, we iteratively perturb the regions (i.e. substitute them with random pixels) in the ascending order of region's saliency score and feed model with these perturbed samples to get the prediction scores, thus we get a curve of prediction score \emph{decay} versus feature perturbation steps. This curve is called LeRF perturbation curve \cite{sanitychecker}, where LeRF is short for Least Relevant First. In this sense, we compute the average of this curve which is denoted as AOPC$_{\texttt{lerf}}$, and the lower value means a less noisy and more faithful saliency map.

Opposite to AOPC$_{\texttt{lerf}}$, the method perturbing those regions with high saliency score first is called MoRF (i.e. Most Relevant First), and we combine these two metrics to a more comprehensive metric called relative AOPC \cite{AOPC} (i.e. AOPC$_{\texttt{rel}}$ $=$ AOPC$_{\texttt{morf}}$ $/$ AOPC$_{\texttt{lerf}}$) to get better estimation of faithfulness for a saliency map.

During evaluation, we perturb a significant part of each test image (20\% regions) in 20 perturbation steps, and each perturbation step is repeated for 5 times. 

\subsection{Main Results}
\begin{figure}[h]
    \centering
    \subfigure[Entropy distributions.]{\includegraphics[width=0.6\columnwidth]{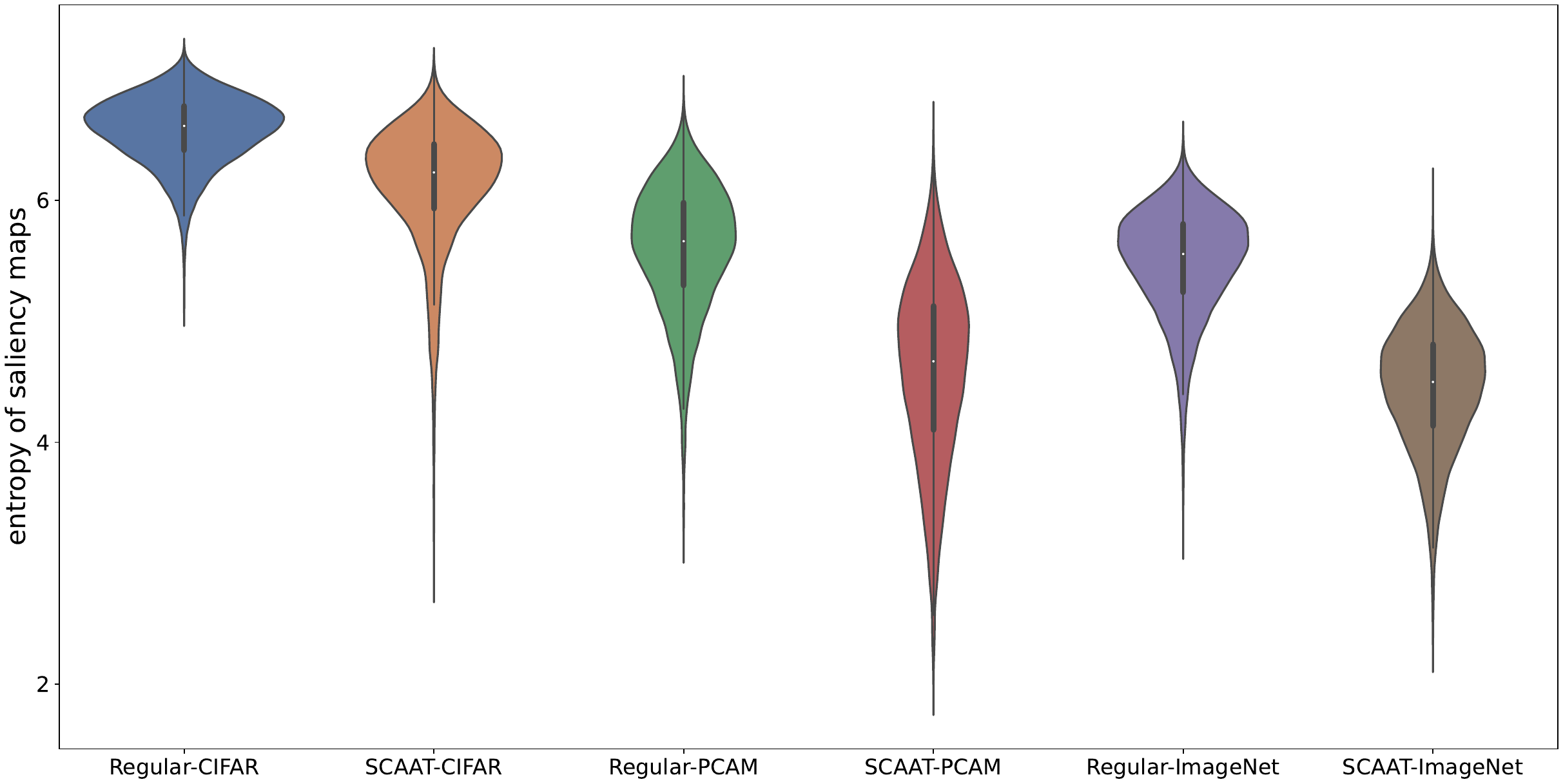}} 
    \hspace{0.00\textwidth}
    \subfigure[Perturbation curves.]{\includegraphics[width=0.3\columnwidth]{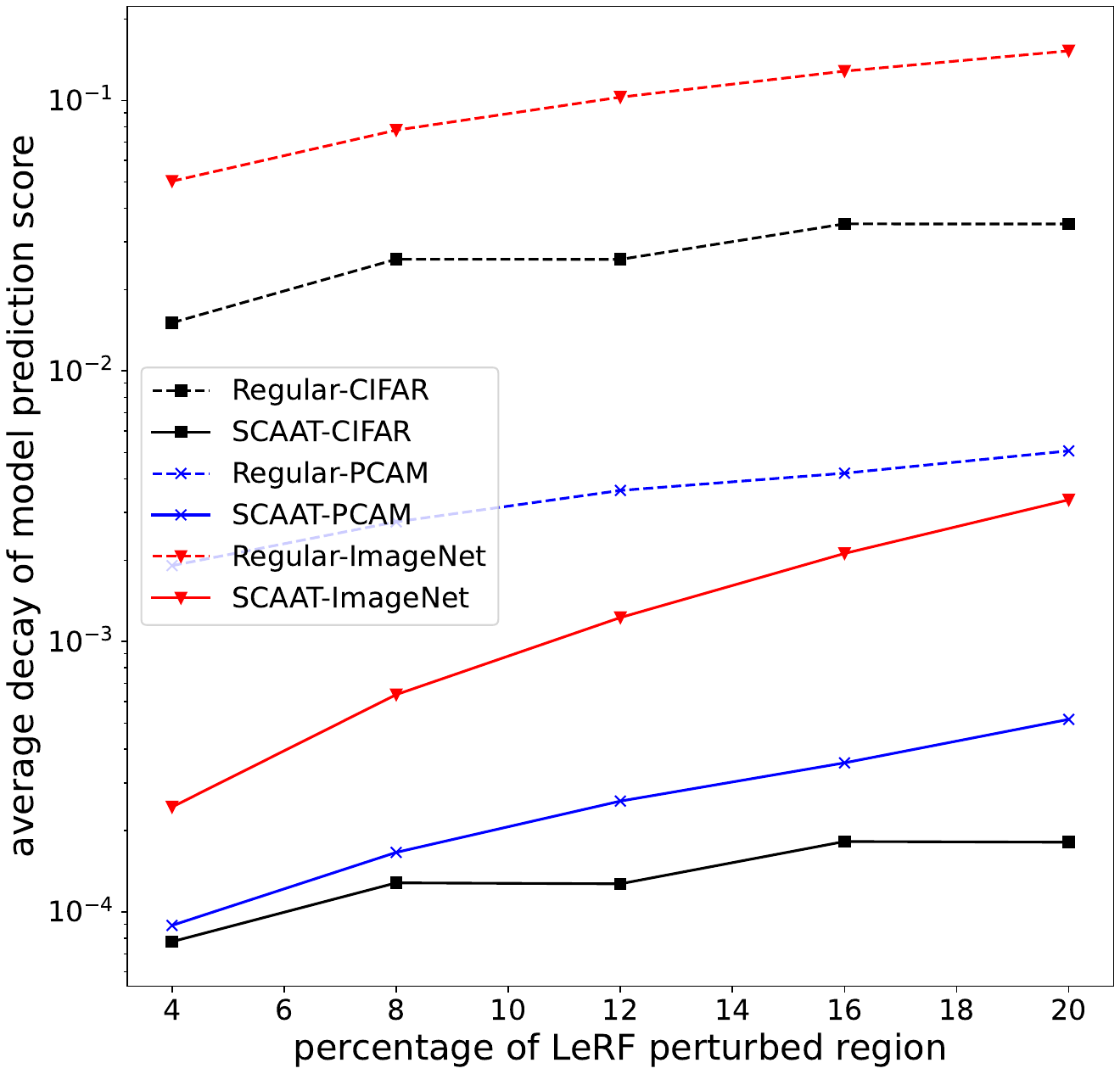}}
    \caption{(a) Comparison of saliency map entropy distributions. (b) Comparison of the confidence-decay curve.}
    \label{fig:foobar}
\end{figure}

\begin{table*}
\caption{\textbf{Evaluation results of saliency map sparsity and faithfulness metrics.} ($\uparrow$) indicates higher numbers are better, while ($\downarrow$) indicates lower numbers are better. We compare our model interpretability with regular trained model (i.e Rgl.) and the model trained with the method proposed by Ismail \etal \cite{ismail2021improving} under different saliency visualization methods. Here the AOPC$_{\texttt{lerf}}$ indicates how sensitive the model is to irrelevant features while AOPC$_{\texttt{rel}}$ is a comprehensive metric for saliency map faithfulness and should be mainly focused. All models are ResNet-18 and the 'C', 'P' and 'I' in the left column stand for the CIFAR-10, PCAM and ImageNet-1k dataset respectively.}
\label{main results}
\begin{center}
\begin{small}
\begin{tabular}{c|l|ccc|ccc|ccc}
\hline
\multirow{2}{*}{} & \multicolumn{1}{c|}{\multirow{2}{*}{Metric}} & \multicolumn{3}{c|}{Vallina Grad}                                        & \multicolumn{3}{c|}{Smooth Grad}                                                        & \multicolumn{3}{c}{Integrated Grad}                                                \\
                         & \multicolumn{1}{c|}{}                        & Rgl.    & Ismail        & Ours                        & Rgl.                        & Ismail           & Ours                        & Rgl.                        & Ismail      & Ours          \\ \hline
\multirow{4}{*}{C}   & Sal. Entropy $\downarrow$                                    & 5.89    & 5.60                        & \textbf{4.69}                        & 5.86                        & 5.69                        & \textbf{4.71}                        & 5.56                        & 5.40                        & \textbf{4.34  }                      \\
                         & Sal. Size (Kbyte) $\downarrow$                                   & 3.30    & 2.93                        & \textbf{1.95}                        & 3.14                        & 2.91                        & \textbf{1.92 }                       & 2.84                        & 2.63                        & \textbf{1.80 }                       \\
                         & AOPC$_{\texttt{lerf}}$$\downarrow$ (\e{-3})                    & 220    & 14.6 & \textbf{0.18} & 180                        & 11.7 & \textbf{0.32 } & 220                        & 38.4 &\textbf{ 0.36} \\
                        & AOPC$_{\texttt{rel}}$ $\uparrow$                                & 2.18    & 17.1                        & \textbf{960}                         & 2.72                        & 22.2                        &\textbf{ 917 }                        & 2.18                        & 6.25                        & \textbf{801    }                 \\    \hline
\multirow{4}{*}{P}    & Sal. Entropy $\downarrow$                                     & 5.61    & 5.30                        & \textbf{ 4.56 }                       & 5.60                        & 5.43                        & \textbf{4.54 }                       & 4.93                        & 4.72                        & \textbf{4.43}                        \\
                         & Sal. Size (Kbyte)  $\downarrow$                                  & 2.48    & 2.34                        & \textbf{1.61 }                       & 2.45                        & 2.33                        & \textbf{1.61  }                      & 2.23                        & 2.04                        &\textbf{ 1.52 }                       \\ 
                         & AOPC$_{\texttt{lerf}}$$\downarrow$ (\e{-3})                               & 3.20 & 6.25 & \textbf{0.23} & 2.89 & 6.29 &\textbf{ 0.23} & 8.94 & 7.65 &\textbf{ 0.21} \\
                         &  AOPC$_{\texttt{rel}}$ $\uparrow$                                & 78.1    & 38.4                        & \textbf{1030 }                       & 90.0                        & 38.1                        & \textbf{982}                         & 24.6                        & 28.8                        & \textbf{938}                         \\ \hline

\multirow{4}{*}{I}    & Sal. Entropy $\downarrow$                                     & 5.49    & 5.21                        & \textbf{ 4.45 }                       & 5.12                       & 5.01                       & \textbf{4.23 }                       & 4.98                        & 4.85                        & \textbf{4.15}                        \\
                         & Sal. Size (Kbyte)  $\downarrow$                                  & 13.2    & 11.9                        & \textbf{7.12 }                       & 12.9                        & 12.7                        & \textbf{6.94  }                      & 12.8                       & 12.4                        &\textbf{ 6.80 }                       \\ 
                         & AOPC$_{\texttt{lerf}}$$\downarrow$ (\e{-3})                               & 85.2 & 42.1 & \textbf{0.98} & 72.5 & 56.9 &\textbf{ 0.93} & 43.2 & 21.8 &\textbf{ 1.21} \\
                         &  AOPC$_{\texttt{rel}}$ $\uparrow$                                & 3.84    & 5.13                        & \textbf{321 }                       & 4.66                        & 6.56                      & \textbf{346}                         & 4.21                        & 7.93                    & \textbf{305}                         \\ \hline
                         
\end{tabular}
\end{small}
\end{center}
\end{table*}
We evaluate the model performance and the quality of our model's saliency map generated by different saliency methods and compare them with that of the baseline and of \cite{ismail2021improving}.
\Cref{main results} shows the comparison of saliency map quality with the baseline and the model proposed by Ismail \etal  \cite{ismail2021improving}. For all of the listed saliency methods and evaluation metrics, our SCAAT beats baseline and Ismail \etal \cite{ismail2021improving} under multiple evaluation metrics for both natural and pathological images. For example, on the ImageNet-1k \cite{deng2009imagenet} dataset, the AOPC$_\texttt{lerf}$ of gradient-based saliency map is decreased from $8.52 \times10^{-2}$ of baseline and $4.21\times10^{-2}$ of \cite{ismail2021improving} to $9.8\times10^{-4}$, which is reduced over one order of magnitude. \Cref{fig:foobar}(a) shows the saliency map entropy distribution of our model and the baseline for all test samples. \Cref{fig:foobar}(b) shows the comparison between the perturbation curves. Specifically, when we perturbs 20\% low-saliency features of the input samples in ImageNet-1k \cite{deng2009imagenet}, the average prediction score decay of baseline model is at the level of \e{-2} while \e{-4} for our SCAAT.

In addition to model interpretability, \Cref{main results perf} compares the performance and interpretability with our method and others. Our method with adaptive $q$ achieves comparable performance to the baseline model and higher than \cite{ismail2021improving} with a significant margin. Specifically, our method performs even better than the baseline model on discrimination power for pathological images.

\begin{figure}[h]
    \centering
    \subfigure[Saliency map comparison.]{\includegraphics[width=0.43\columnwidth]{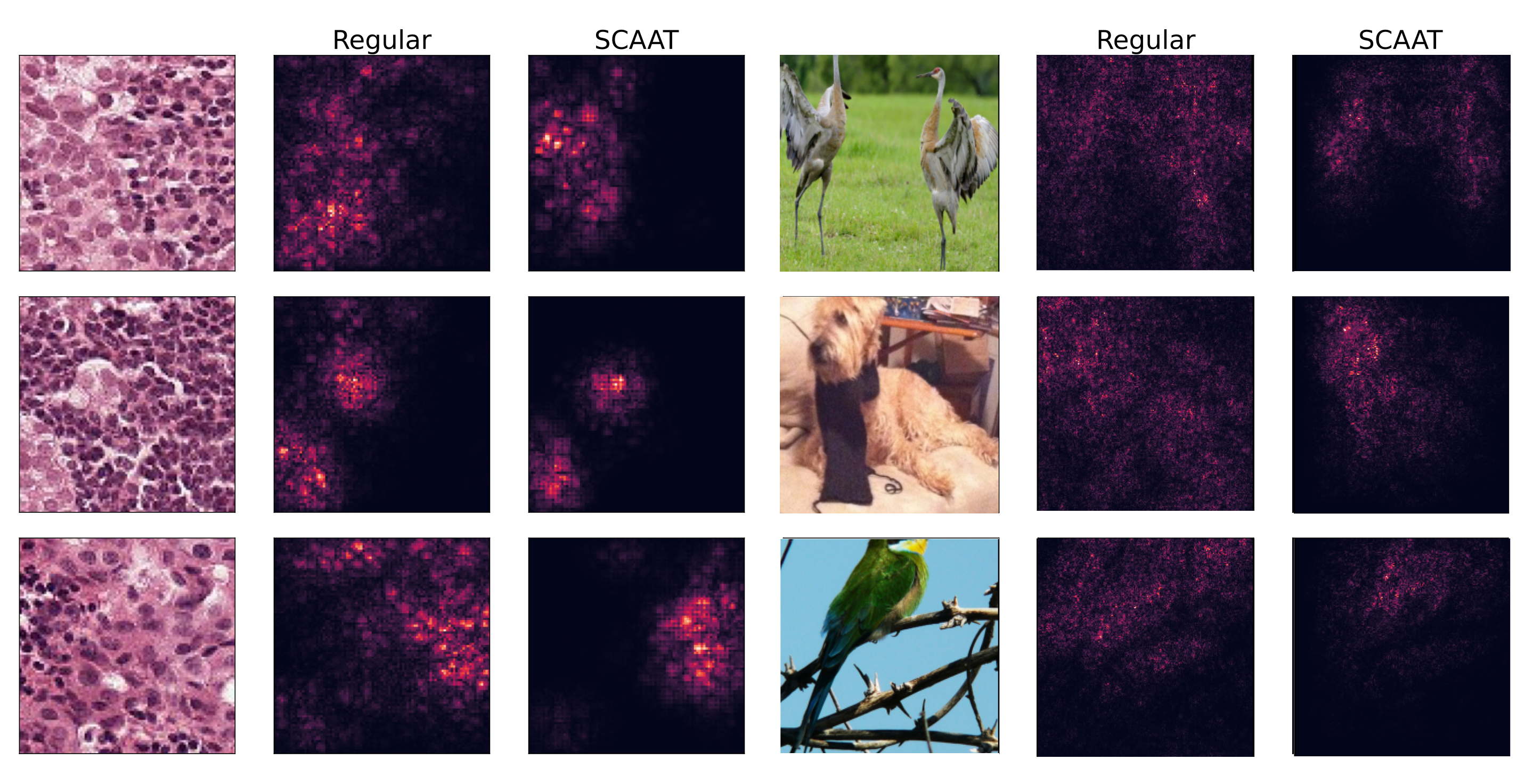}} 
    \hspace{0.00\textwidth}
    \subfigure[Samples of different $q$ values.]{\includegraphics[width=0.4\columnwidth]{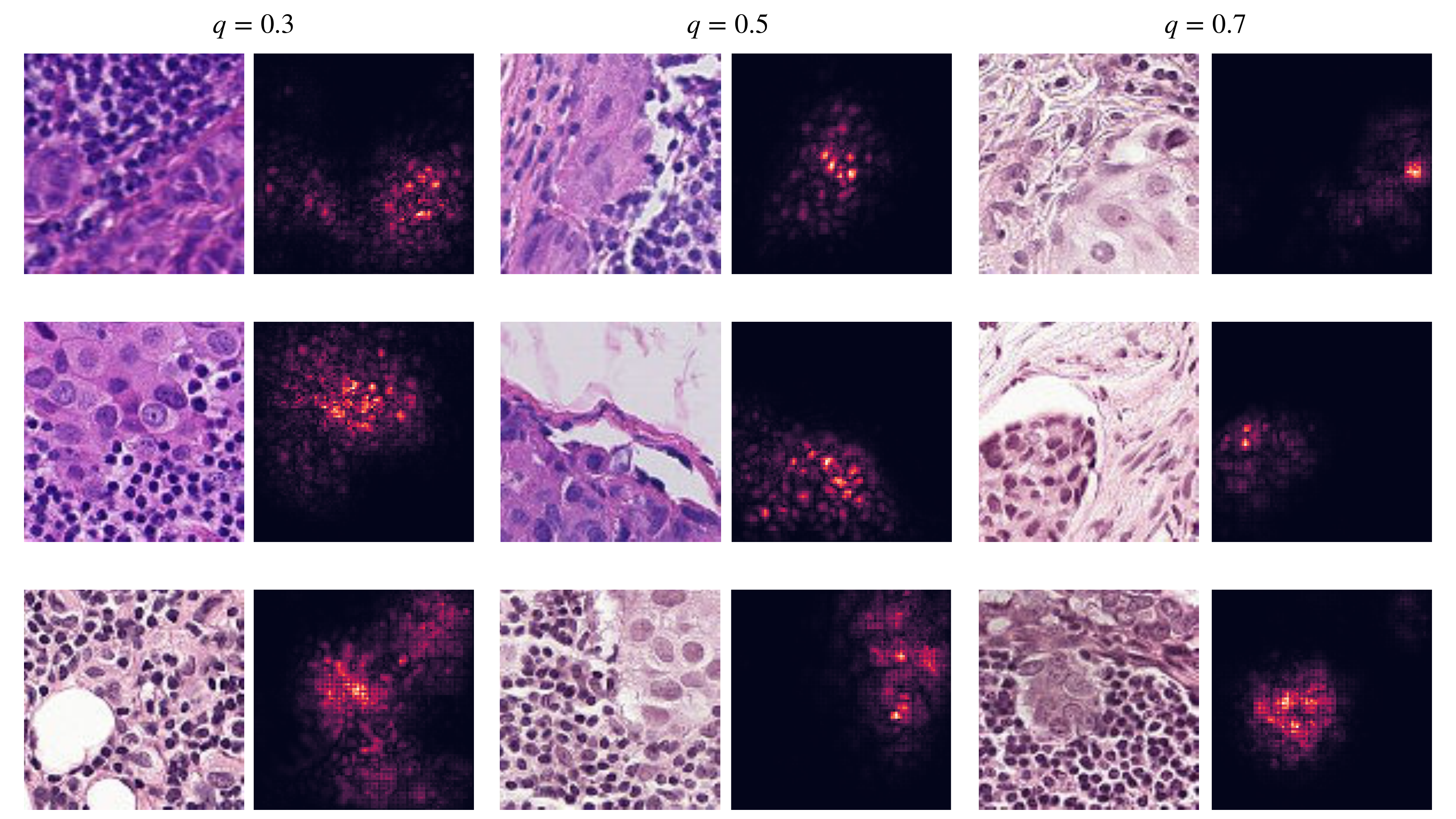}}
    \caption{(a) Visualizations of saliency maps of baseline model and ours. Left pane is for PCAM and right pane is for ImageNet-1k. (b) Visualizations of training samples and their saliency maps with values of $q$ which are searched in the training process.}
    \label{fig:saliency maps and adaptive q}
\end{figure}

\Cref{fig:saliency maps and adaptive q}(a) visualizes the saliency map of the ResNet-18 \cite{resnet} model trained by SCAAT and baseline method on both PCAM \cite{pcam} and ImageNet-1k \cite{deng2009imagenet} dataset. Obviously our method enhances the model's sensitivity to critical features while suppressing the noise on irrelevant features, so the saliency map looks more sparse and accurately indicates the critical features. \Cref{fig:saliency maps and adaptive q}(b) shows the representative training samples in PCAM \cite{pcam} of different $q$ which are adjusted during the process of our adaptive adversarial training. Obviously the images with more uncritical regions will be assigned larger $q$ values adaptively in the training process, which means we can perturb more their irrelevant features to get cleaner saliency map without making the model misclassify them.

\begin{table}
\caption{\textbf{Performance comparison.} We test the performance of baseline methods and ours under fixed or adaptive saliency constraint. The metric of performance is top-1 ACC for CIFAR-10 \cite{cifar} and ImageNet-1k \cite{deng2009imagenet} while AUC for PCAM \cite{pcam}, and Intp. indicates the AOPC$_{\texttt{rel}}$ of gradient-based saliency map. The models trained by our method have comparable performance with baseline while the interpretability is significantly improved.}
\label{main results perf}
\begin{center}
\begin{small}
\begin{tabular}{c|l|cc|cc}
\hline
\multirow{2}{*}{Dataset}  & \multicolumn{1}{c|}{\multirow{2}{*}{Method}} & \multicolumn{2}{c|}{ResNet-18} & \multicolumn{2}{c}{VGG-16} \\
                          & \multicolumn{1}{c|}{}                        & Perf.      & Intp.     & Perf.     & Intp.    \\ \hline
\multirow{4}{*}{CIFAR-10} & Regular                                      & \textbf{0.910}          & 2.18               & \textbf{0.904}         & 2.35               \\
                          & Ismail                            & 0.892             & 17.1               & 0.889         & 15.2               \\
                          & Ours (fixed $q$)                            & 0.890              & 871                & 0.893         & 896               \\
                          &  Ours (adpt. $q$)                         & 0.905          & \textbf{960}                       & 0.901         & \textbf{921}               \\ \hline
\multirow{4}{*}{PCAM}     & Regular                                      & 0.928          & 78.1                  & 0.935           & 85.3               \\
                          & Ismail                             &  0.911          &  38.4               & 0.929            & 78.4               \\
                          & Ours (fixed $q$)                            & 0.926           & 987                   & 0.931            & 801               \\
                          &  Ours (adpt. $q$)                         & \textbf{0.933}          & \textbf{1030 }                     & \textbf{0.939 }           & \textbf{956}              \\ \hline
\multirow{4}{*}{ImageNet-1k}     & Regular                                      & \textbf{0.687 }        & 3.84                & \textbf{0.744}           & 4.29               \\
                          & Ismail                             &  0.653          &  5.13               & 0.684            & 6.05               \\
                          & Ours (fixed $q$)                            & 0.671           & 215                   & 0.721           & 266               \\
                          &  Ours (adpt. $q$)                         & 0.682         & \textbf{321}                     & 0.738           & \textbf{368}              \\ \hline

\end{tabular}
\end{small}
\end{center}
\vskip -0.1in
\end{table}

Adaptive adversarial training is the core module of SCAAT to improve the model's interpretability. Thus we did a lot of  experiments for the adaptive $q$-selecting algorithm, the loss function of divergence, and the searching radius for perturbations. These detailed results are shown in the supplementary materials.

\subsection{Training Efficiency}
The extra training cost for our SCAAT mainly comes from adversarial sample generation, which just requires several back-propagation steps and depends on the searching algorithm. The default step is set to 4 in our work, which makes the training time about 2.5$\times$ longer than regular training. Additionally, FGSM\cite{goodfellow2014explaining} fast searching requires only one extra gradient step, leading to slight interpretability sacrifice but significant efficiency gains (see the last row in \Cref{table:pgd ablative}). There is a trade-off between quality of adversarial samples and the computational efficiency.

Towards the computational efficiency, we simply determine irrelevant features in each image according to the saliency map of vallina gradients. Despite requiring several extra back-propagation steps, we also experimented with Smooth Grad \cite{smilkov2017smoothgrad}, which more precisely indicates uncritical features then prevents the model from being desensitised to the critical features. The performance can be further improved by introducing this more advanced saliency method, but the computational overhead will be greatly increased during training.

\begin{table}[H]
\caption{\textbf{Efficiency Comparison.} We evaluate the extra training cost of our method. The Gini Index proposed in \cite{chalasani2020concise} measures the sparsity of saliency maps.}
\begin{center}
\begin{small}
\begin{tabular}{c|c|c|c|c}
\hline
Dataset                   & Method  & Acc(\%)   & Gini Index $\uparrow$ & Time \\ \hline
\multirow{3}{*}{ImageNet-1k} & Regular & 68.7$\pm$0.1 & 0.455  & 1.0$\times$  \\
                          & Ours (PGD-4)   & 68.3$\pm$0.2 & 0.601 & 2.5$\times$ \\ 
                          & Ours (FGSM)   & 68.4$\pm$0.2 & 0.578  &  1.4$\times$\\ \hline
\end{tabular}
\end{small}
\end{center}
\label{table:pgd ablative}
\end{table}

\section{Conclusion}
In this work, we focus on improving interpretability of deep neural networks by denoising the saliency maps of a model. We proposed a model-agnostic adversary-based training method using saliency map as constraints to desensitize a model to irrelevant features. Motivated by the observation that the ratio of irrelevant features varies across training samples, the proposed method iteratively estimates the ratio of irrelevant features in a saliency map for further desensitizing perturbation, according to the dynamic impact on the model. Experiments showed our proposed training method achieves significant improvement on the quality of saliency map for both natural and pathological images without sacrifying model performance.

\section{Acknowledgement}
This research was supported in part by the NIFDC Key Technology Research Grant (GJJS-2022-3-1). We thank Qiuchuan Liang for doing some data processing work.
\bibliography{egbib}
\end{document}